\documentclass[letterpaper, 10 pt, journal, twoside]{ieeetran}
\usepackage[utf8]{inputenc} 
\usepackage[T1]{fontenc}    
\usepackage{hyperref}       
\usepackage{url}            
\usepackage{booktabs}       
\usepackage{amsfonts}       
\usepackage{nicefrac}       
\usepackage{microtype}      
\usepackage{xcolor}         

\usepackage{bm}
\usepackage{amsmath}
\usepackage{amssymb}
\usepackage{mathtools}
\usepackage{amsthm}
\usepackage{caption}
\usepackage{multirow}
\usepackage{makecell}
\usepackage{fontawesome}
\usepackage{marvosym}
\usepackage[top=57pt, bottom=43pt, left=48pt,right=48pt]{geometry}
\usepackage{graphicx}
\usepackage{wrapfig}
\usepackage{subcaption}

\usepackage[ruled,vlined]{algorithm2e}
\usepackage{algorithmicx}
\usepackage{algpseudocode}
\usepackage{makecell}

\begin{document}

\title{What Matters in Learning A Zero-Shot Sim-to-Real RL Policy\\ for Quadrotor Control? A Comprehensive Study}

\author{Jiayu Chen$^{1*}$, Chao Yu$^{12*\textsuperscript{\Letter}}$, Yuqing Xie$^{1}$, Feng Gao$^{1}$, Yinuo Chen$^{1}$, Shu'ang Yu$^{13}$, \\ Wenhao Tang$^{4}$, Shilong Ji$^{1}$, Mo Mu$^{1}$, Yi Wu$^{1}$, Huazhong Yang$^{1}$, Yu Wang$^{1\textsuperscript{\Letter}}$
\thanks{Manuscript received: December, 17, 2024; Revised March, 25, 2025; Accepted May, 20, 2025.}
\thanks{This paper was recommended for publication by Editor Aleksandra Faust upon evaluation of the Associate Editor and Reviewers' comments.}
\thanks{This research was supported by National Natural Science Foundation of China (No.62406159, 62325405), Postdoctoral Fellowship Program of CPSF under Grant Number (GZC20240830, 2024M761676), China Postdoctoral Science Special Foundation 2024T170496.}
\thanks{* Equal Contribution. {\Letter} Corresponding Authors. \url{zoeyuchao@gmail.com} and \url{yu-wang@tsinghua.edu.cn}}
\thanks{$^{1}$Tsinghua University, Beijing, 100084, China. $^{2}$Beijing Zhongguancun Academy,
Beijing, 100094, China. $^{3}$Shanghai Artificial Intelligence Laboratory,
Shanghai, 200030, China. $^{4}$Tsinghua Shenzhen International Graduate School,
Shenzhen, 518055, China.}
\thanks{Digital Object Identifier (DOI): see top of this page.}}

\maketitle

\begin{abstract}
{Precise and agile flight maneuvers are essential for quadrotor applications, yet traditional control methods are limited by their reliance on flat trajectories or computationally intensive optimization. Reinforcement learning (RL)-based policies offer a promising alternative by directly mapping observations to actions, reducing dependency on system knowledge and actuation constraints. However, the sim-to-real gap remains a significant challenge, often causing instability in real-world deployments.  In this work, we identify five key factors for learning robust RL-based control policies capable of zero-shot real-world deployment: (1) integrating velocity and rotation matrix into actor inputs, (2) incorporating time vector into critic inputs, (3) regularizing action differences for smoothness, (4) applying system identification with selective randomization, and (5) using large batch sizes during training. Based on these insights, we develop \textit{SimpleFlight}, a PPO-based framework that integrates these techniques.  Extensive experiments on the Crazyflie quadrotor demonstrate that SimpleFlight reduces trajectory tracking error by over 50\% compared to state-of-the-art RL baselines. It excels in both smooth polynomial and challenging infeasible zigzag trajectories, particularly on small thrust-to-weight quadrotors, where baseline methods often fail. To enhance reproducibility and further research, we integrate SimpleFlight into the GPU-based Omnidrones simulator and provide open-source code and model checkpoints. For more details, visit our project website at \url{https://sites.google.com/view/simpleflight/}.}
\end{abstract}

\section{INTRODUCTION}\label{sec:intro}
\IEEEPARstart{P}{recise} and agile flight maneuvers are essential for UAVs, especially quadrotors, in applications such as package delivery~\cite{grzybowski2020low}, search and rescue~\cite{scherer2015autonomous}, and infrastructure inspection~\cite{nikolic2013uav}. Traditional control methods, whether model-based or model-free, often face limitations due to their dependence on flat trajectories adhering to actuation constraints~\cite{5980409,8118153} or the need for accurate system modeling and nonconvex optimization solvers~\cite{hanover2021performance,7989202}, which can restrict policy flexibility. Recently, reinforcement learning (RL) has gained traction as a versatile and efficient alternative for quadrotor control~\cite{hwangbo2017control,kiumarsi2017optimal}. RL-based policies map observations directly to actions, bypassing the need for actuation constraints or precise system dynamics knowledge~\cite{pfeiffer2022visual}, enabling lower control latency and the potential for enhanced performance in quadrotor tasks.

A major challenge in RL-based quadrotor control is the sim-to-real gap, where policies trained in simulation often fail to perform stably in real-world deployment without additional fine-tuning. Despite numerous RL-based approaches, there is no unified understanding of the key factors for training robust, zero-shot deployable policies~\cite{huang2023datt,gao2024neural,song2021autonomous,10517383,kaufmann2022benchmark,dionigi2024power}. For example, while reward functions are often designed to constrain control commands and enhance smoothness, the specific reward components critical for ensuring valid commands and task success remain unclear. domain randomization is widely employed to bridge the sim-to-real gap, but the extent to which it is necessary for effective quadrotor control remains unclear. Additionally, other factors influencing the training of robust RL-based policies remain underexplored.

This study explores crucial factors for developing robust RL-based control policies for real-world zero-shot deployment. We identify five key elements across input space design, reward design, system identification and training techniques: (1) integrating velocity and rotation matrix into actor's input, (2) adding time vector to critic's input, (3) employing action difference regularization as smoothness reward, (4) applying system identification to key dynamics parameters with selective randomization, and (5) utilizing large batch sizes. The first two factors optimize input space design for simulation learning, while the remaining three work together to bridge the sim-to-real gap. We implement these techniques in a PPO-based framework, \textbf{\textit{SimpleFlight}}.

Our experimental validation on the Crazyflie 2.1 nano quadrotor demonstrates SimpleFlight's superior performance. The framework achieves over 50\% reduction in trajectory tracking error compared to SOTA RL baselines without specialized algorithmic or architectural modifications. SimpleFlight successfully completes all benchmark trajectories, including challenging infeasible trajectories, outperforming other RL policies that fail under extreme conditions. {Moreover, SimpleFlight demonstrates consistent performance across different quadrotor platforms, including the Crazyflie and our custom-built quadrotor, showcasing its strong generalization capability across varying models and sizes.}

Furthermore, we integrate SimpleFlight into a high-parallel GPU-based simulator Omnidrones~\cite{xu2024omnidrones}, and we open-source the code, model checkpoints, and benchmark tasks to ensure reproducibility. We believe that SimpleFlight will provide valuable insights to guide future research in RL-based quadrotor control. Our contributions can be summarized as follows:

\begin{itemize}
\item We investigate five key learning factors and develop a PPO-based training framework, SimpleFlight, for learning RL-based zero-shot sim-to-real policies.
\item We conduct extensive real-world experiments on the Crazyflie to demonstrate the effectiveness of SimpleFlight. The policy derived by SimpleFlight is the only one capable of successfully completing all benchmarking trajectories, including both smooth and infeasible trajectories.
\item SimpleFlight reduces trajectory tracking error by over 50\% compared to SOTA RL baselines, despite not employing any tailored algorithmic or network architecture design.
\item We integrate SimpleFlight into the high-parallel GPU-based simulator Omnidrones and open-source checkpoints to ensure reproducibility. 
\end{itemize}
\begin{figure*}[t]
  \centering
\includegraphics[width=0.78\textwidth]{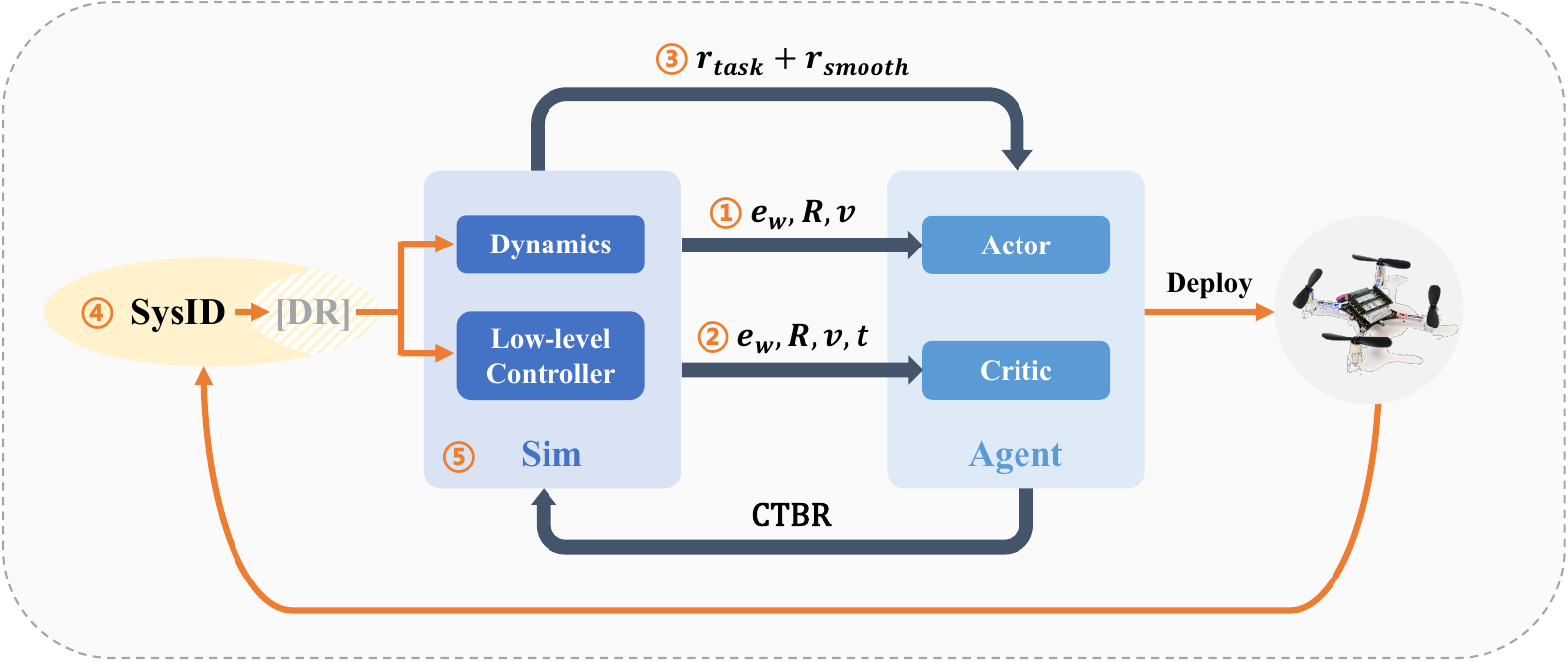}
  \caption{Overview of SimpleFlight. We begin with SysID and selective DR for quadrotor dynamics and low-level control. Next, an RL policy is trained in simulation to output CTBR for tracking arbitrary trajectories and zero-shot deployed directly on a real quadrotor. The training framework focuses on three key aspects, i.e., input space design, reward design, system identification and domain randomization, as well as training techniques, identifying five critical factors to enhance zero-shot deployment.}
  \label{fig:overview}
  \vspace{-5mm}
\end{figure*}
\vspace{-1mm}

\section{RELATED WORK}\label{sec:related}
{\subsection{General Approaches for Sim-to-Real Transfer in Robotics}
Bridging the sim-to-real gap is a critical challenge in robotics, with system identification (SysID) being one of the most straightforward approaches~\cite{kristinsson1992system}. While simulators cannot fully replicate real-world complexities, constructing accurate mathematical models within the simulator can significantly improve the real-world performance of reinforcement learning (RL) policies. However, the dynamic and time-varying nature of the real world, combined with factors such as friction and motor delays, makes strict calibration of simulator parameters through SysID challenging. To address these limitations, domain randomization (DR) has emerged as a promising technique~\cite{tobin2017domain, tremblay2018training}. DR randomizes simulator parameters to cover a broader range of real-world conditions, thereby enhancing policy robustness during deployment.

DR techniques can be categorized into two types: dynamics randomization and visual randomization. Dynamics randomization, particularly effective for control tasks, improves policy stability by randomizing physical parameters such as object size, friction, mass, and damping coefficients~\cite{andrychowicz2020learning, molchanov2019sim}. For example, in OpenAI's work on solving a Rubik's Cube using a robotic hand, randomization of these parameters enabled the policy to adapt to diverse real-world conditions. Visual randomization, on the other hand, is primarily used in vision-based tasks to address discrepancies in textures, lighting, and camera perspectives~\cite{tobin2017domain, loquercio2019deep}.

In addition to DR, domain adaptation techniques have been widely adopted to minimize the distributional discrepancy between simulation (source domain) and reality (target domain)~\cite{ganin2016domain,bousmalis2017unsupervised, jing2023unsupervised, long2015learning, park2021sim}. These techniques are particularly effective for vision-based robotics tasks. In this work, we focus on evaluating the impact of SysID and dynamics randomization on sim-to-real performance, leaving visual randomization and domain adaptation for future exploration.

\subsection{Sim-to-Real approaches for Quadrotors}
In the field of quadrotors, significant advancements have been made in SysID and DR to bridge the sim2real gap. For SysID, Gronauer et al.~\cite{gronauer2022using} employ Bayesian optimization to automatically calibrate the simulator's dynamic parameters using real-world flight data. Bauersfeld et al.~\cite{bauersfeld2021neurobem, kaufmann2023champion} utilize residual models and real flight data to accurately model the aerodynamics of drones during high-speed flight, further enhancing the fidelity of simulation environments.

For DR, Molchanov et al.~\cite{molchanov2019sim} randomize key dynamic parameters of the drone, enabling successful policy transfer from simulation to various types of quadrotors. Zhang et al.~\cite{zhang2023learning} combine adaptive control with dynamic parameter randomization, transferring policy to real-world environments with unknown disturbances. Additionally, some studies focus on improving RL policy performance in real-world scenarios by designing specialized observation spaces~\cite{dionigi2024power}, action spaces~\cite{kaufmann2022benchmark}, and incorporating supplementary training methods~\cite{eschmann2024learning}.

Despite these advancements, several critical challenges remain unresolved. First, there is no unified framework or standardized pipeline for training robust, zero-shot deployable RL policies. Existing approaches often rely on ad hoc designs for observation spaces, reward functions, and training strategies, making it difficult to generalize findings across different platforms and tasks. Second, while SysID and DR are widely used, their individual and combined effects on sim2real performance are not well understood. For instance, it remains unclear which dynamic parameters are most sensitive to SysID or DR, and under what conditions DR may hinder rather than improve performance. Third, many studies focus on specific aspects of the sim2real pipeline (e.g., observation design or parameter calibration) without considering their interplay, leading to suboptimal solutions. Finally, there is a lack of systematic evaluation of training strategies, such as batch size optimization, which can significantly impact sim2real performance without requiring additional modifications.}
\vspace{-1mm}

\section{PRELIMINARY}\label{sec:prelim}
\subsection{Problem Formulation}
We formulate the quadrotor control problem as a Markov Decision Process (MDP). The MDP is defined as $M = <\mathcal{S}, \mathcal{A}, \mathcal{O}, P, R, \gamma>$, with the state space $\mathcal{S}$, the action space $\mathcal{A}$, the observation space $\mathcal{O}$, the transition probability $\mathcal{P}$, the reward function $\mathcal{R}$ and the discount factor $\gamma$. Denote the state and the observation at time step $t$ as $(s_t, o_t)\in (\mathcal{S}, \mathcal{O})$. The goal of our work is to construct a policy $\pi_\theta$ parameterized by $\theta$ to output action $a_t \sim \pi_\theta(o_t)$ that performs precise and agile maneuvers to\textit{ track arbitrary trajectories}. The optimization objective is to maximize the expected accumulative reward $J(\theta)= \mathbb{E} \left[\sum_t \gamma^t R(s_t, a_t)\right]$.

\subsection{Quadrotor Dynamics}
The quadrotor is assumed to be a $6$ degree-of-freedom rigid body of mass $m$ and diagonal moment of inertia matrix $\textbf{\textit{I}} = diag(I_x, I_y, I_z)$. 
The state space is 17-dimensinal and the dynamics are modeled by the differential equation: 
\begin{equation}
    \boldsymbol{\dot{x}} = 
    \begin{bmatrix}
    \boldsymbol{\dot{p}}_{\mathcal{W}} \\
    \boldsymbol{\dot{q}} \\
    \boldsymbol{\dot{v}}_{\mathcal{W}} \\
    \boldsymbol{\dot{\omega}}_{\mathcal{B}}\\
    
    \boldsymbol{\dot{\Omega}}
    \end{bmatrix}
    = 
    \begin{bmatrix}
    \boldsymbol{v}_{\mathcal{W}} \\
    \boldsymbol{q}\textcircled{\boldmath$\times$}[0, \boldsymbol{\omega}_{\mathcal{B}}/2]^T\\
    \frac{1}{m}\boldsymbol{q}\cdot\boldsymbol{f}_{prop}\cdot\overline{\boldsymbol{q}} + \boldsymbol{g}_{\mathcal{W}} \\
    \boldsymbol{I}^{-1}(\boldsymbol{\tau}_{prop} - \boldsymbol{\omega}_{\mathcal{B}}\times(\boldsymbol{I}\boldsymbol{\omega}_{\mathcal{B}})) \\
    T_m(\boldsymbol{\Omega}_{cmd} - \boldsymbol{\Omega})
    \end{bmatrix},
\end{equation}
where the quadrotor state $\boldsymbol{x}$ consists the position $\boldsymbol{p}$, the orientation $\boldsymbol{q}$ in quaternions, the linear velocity $\boldsymbol{v}$, the angular velocity $\boldsymbol{\omega}$ and the rotational speed of the rotor $\boldsymbol{\Omega}$. $m$ denotes mass. The subscripts $\mathcal{W}$ and $\mathcal{B}$ represent the world and body frame. The frame $\mathcal{B}$ is located at the center of the mass of the quadrotor. The notation $\textcircled{\boldmath$\times$} $ indicates the multiplication of two quaternions. $\overline{\boldsymbol q}$ is the quaternion’s conjugate. $\boldsymbol{g}_{\mathcal{W}} = [0, 0, -9.81m/s^2]^T$ denotes earth's gravity. $\boldsymbol{f}_{prop}$ and $\boldsymbol{\tau}_{prop}$ are the collective force and the torque produced by the propellers. The quantities are defined as follows,
\begin{equation}
    \boldsymbol{f}_{prop} = \Sigma_i\boldsymbol{f}_j, \boldsymbol{\tau}_{prop} = \Sigma_j\boldsymbol{\tau}_j + \boldsymbol{r}_{p, j}\times\boldsymbol{f}_j,
\end{equation}
where $\boldsymbol{r}_{p, j}$ is the location of propeller $j$ expressed in the body frame, $\boldsymbol{f}_j, \boldsymbol{\tau}_j$ are the forces and torques generated by the $j$-th propeller. The rotational speeds of the four motors ${\Omega}_j$ are modeled as a first-order system with a time constant $T_m$, where the commanded rotational speeds $\boldsymbol{\Omega}_{cmd}$ serve as the input. For the forces and torques generated by each motor, we adopt a widely used model from prior work~\cite{furrer2016rotors, shah2018airsim}:
\begin{equation}
    \boldsymbol{f}_j = [0, 0, k_f\Omega_j^2]^T, \boldsymbol{\tau}_j = [0, 0, k_m\Omega_j^2]^T.
\end{equation}
The thrust and torque are modeled as proportional to the square of the motor's rotational speed. The corresponding thrust and drag coefficients, $k_f$ and $k_m$, as well as the motor time constant, $T_m$, can be determined using a static propeller test stand.

\section{SIMPLEFLIGHT}\label{sec:method} 
\subsection{Overview}
In this section, we describe the details of the entire training framework SimpleFlight, as shown in Fig.~\ref{fig:overview}. The core idea of learning a zero-shot sim-to-real RL policy involves two main aspects: improving policy performance in simulation and bridging the sim-to-real gap to minimize performance drop when deploying the policy in the real world. 

To this end, we follow standard practices by first performing SysID to bridge the sim-to-real gap. We calibrate the quadrotor's dynamics by estimating four key parameters: mass $m$, inertia matrix $\boldsymbol{I}$, thrust coefficient $k_f$, and motor time constant $T_m$. These calibrated parameters are then used to model the quadrotor in the simulation.
Following previous work, we adopt CTBR as the policy action space, a mid-level control command that has been shown to be more robust to sim-to-real gap~\cite{kaufmann2022benchmark}. To convert CTBR commands into four motor thrusts, a low-level controller is employed. Furthermore, we align the low-level controller in the quadrotor firmware with the one used in the simulator by calibrating its system response, thereby further bridging the sim-to-real gap.

Next, we train an RL policy in a simulator to enable the quadrotor to track arbitrary trajectories, which is then directly deployed on a real quadrotor without fine-tuning. The training framework emphasizes four critical aspects to enhance zero-shot deployment performance: input space design, reward design, system identification and domain randomization, as well as training techniques. Specifically, input space design aims to improve policy performance in simulation, while the other techniques are tailored to reduce the sim-to-real gap. From these three aspects, we identify and summarize five critical factors. The final design of SimpleFlight is detailed in the following sections, while comparisons with alternative configurations are presented in the experiment section.

\subsection{Input Space Design}
We employ a custom asymmetric actor-critic architecture in SimpleFlight. The actor takes as input the relative positions of the quadrotor to the next $N$ reference trajectory points in the world frame $\boldsymbol{e}_{\mathcal{W}}$, the linear velocity $\boldsymbol{v}$, and the rotation matrix $\boldsymbol{R}$. This design allows the policy to perform long-horizon planning, which is particularly crucial for tracking infeasible trajectories with sharp turns. In practice, we set $N=10$, with each point spaced by $0.05$s. The critic, on the other hand, receives the same inputs as the actor, augmented with a time vector $\mathbf{f}_t=[t,...,t]^T\in \mathbb{R}^k$ as privileged information, {where $t$ denotes the discrete timestep in RL training. Since the training performance is not sensitive to the choice of $k$, we simply set $k = 1$ in our task. While for more complex tasks with high-dimensional observations, we recommend to increase the dimension $k$ if the value estimation proves insufficiently accurate.} The time vector enables the critic to capture temporal information, improving its ability to estimate state values effectively. {Both the actor and critic use three-layer MLPs, with an ELU activation function and a LayerNorm layer inserted between the two MLP layers to encode their inputs into a latent vector, respectively. The output dimension of the MLP layer is 256.} For the actor, this vector parameterizes a Gaussian distribution to generate CTBR actions. For the critic, the vector is further fed into an MLP to produce the estimated state value. 

\textbf{Factor 1}: Utilizing the rotation matrix instead of a quaternion and incorporating linear velocity into the actor's input.

\textbf{Factor 2}: Adding a time vector to the critic's input to enhance its temporal awareness.

\subsection{Reward Design}
In simulation, RL policies often explore aggressive actions to optimize task performance. For instance, a policy might produce a collective thrust command that abruptly changes from maximum thrust at one timestep to 0 at the next. While such actions may execute without immediate instability in simulation, due to the sim-to-real gap, they become physically infeasible for real quadrotors and can lead to unstable behavior during deployment.

To address this, RL leverages reward design to regularize policy outputs. Existing studies incorporate auxiliary reward components to encourage smooth actions. In general, the reward function is defined as:
\begin{equation}
    r = r_{task} + \lambda r_{smooth},
\end{equation}
where $r_{task}$ represents the task-specific reward, and 
$r_{smooth}$ is a smoothness reward designed to promote smooth actions. $\lambda$ is the coefficient of the smoothness reward. We normalize both $r_{task}$ and $r_{smooth}$ to the range $[0, 1]$, allowing $\lambda$ to represent the relative weight of the smoothness reward compared to the task-specific reward. 

It is important to note the trade-off between $r_{task}$ and $r_{smooth}$. While $r_{smooth}$ discourages aggressive actions and enforces smoother commands, it can also restrict the quadrotor's ability to exploit agile maneuvers essential for tackling complex and challenging tasks. In SimpleFlight, we adopt the form $||\boldsymbol{u}_t - \boldsymbol{u}_{t-1}||_2$ as the smoothness reward, where $\boldsymbol{u}_t$ represents the policy's action, i.e., CTBR. We hypothesize that this design penalizes abrupt changes in policy output, serving as a soft yet direct constraint. This approach achieves a better trade-off between task completeness and action smoothness, enabling both stable and agile flight.

\textbf{Factor 3}: Incorporating regularization of the difference between successive actions as the smoothness reward.

{
\subsection{System Identification and Domain Randomization}
Our study explores the effects of system identification (SysID), domain randomization (DR), and their combination on sim2real performance with four key dynamic parameters, i.e., mass $m$, inertia $\boldsymbol{I}$, motor time constant $T_m$, and thrust coefficient $k_f$. We find that SysID is crucial for accurate sim2real transfer, particularly for measurable parameters like mass $m$ and inertia $\boldsymbol{I}$, where DR is counterproductive, increasing training complexity and leading to suboptimal policies. For the motor time constant $T_m$, sim2real performance shows low sensitivity, and DR provides no significant benefits, only introducing unnecessary learning challenges. In contrast, the thrust coefficient $k_f$ is highly sensitive, with deviations causing notable performance drops. However, DR significantly improves robustness.

\textbf{Factor 4}: Applying SysID for calibrating key dynamic parameters is crucial. DR exhibits selective effectiveness, improving performance for sensitive parameters like thrust coefficients while proving detrimental for insensitive or accurately measurable parameters.

\subsection{Training Techniques}
We highlight a often-overlooked training technique that significantly impact policy performance. Increasing the batch size during training without requiring additional modifications improves real-world performance despite having limited impact on simulation results. This benefit possibly arises from the enhanced data diversity generated by larger batch sizes, which strengthens the policy’s generalization capacity.

\textbf{Factor 5}: Leveraging larger batch sizes during training.}

\section{EXPERIMENT}\label{sec:expr}
\begin{figure*}[t]
\centering
\begin{minipage}{0.99\textwidth}
\centering
\subcaptionbox{Figure-eight (fast).}
{\includegraphics[width=0.25\textwidth]{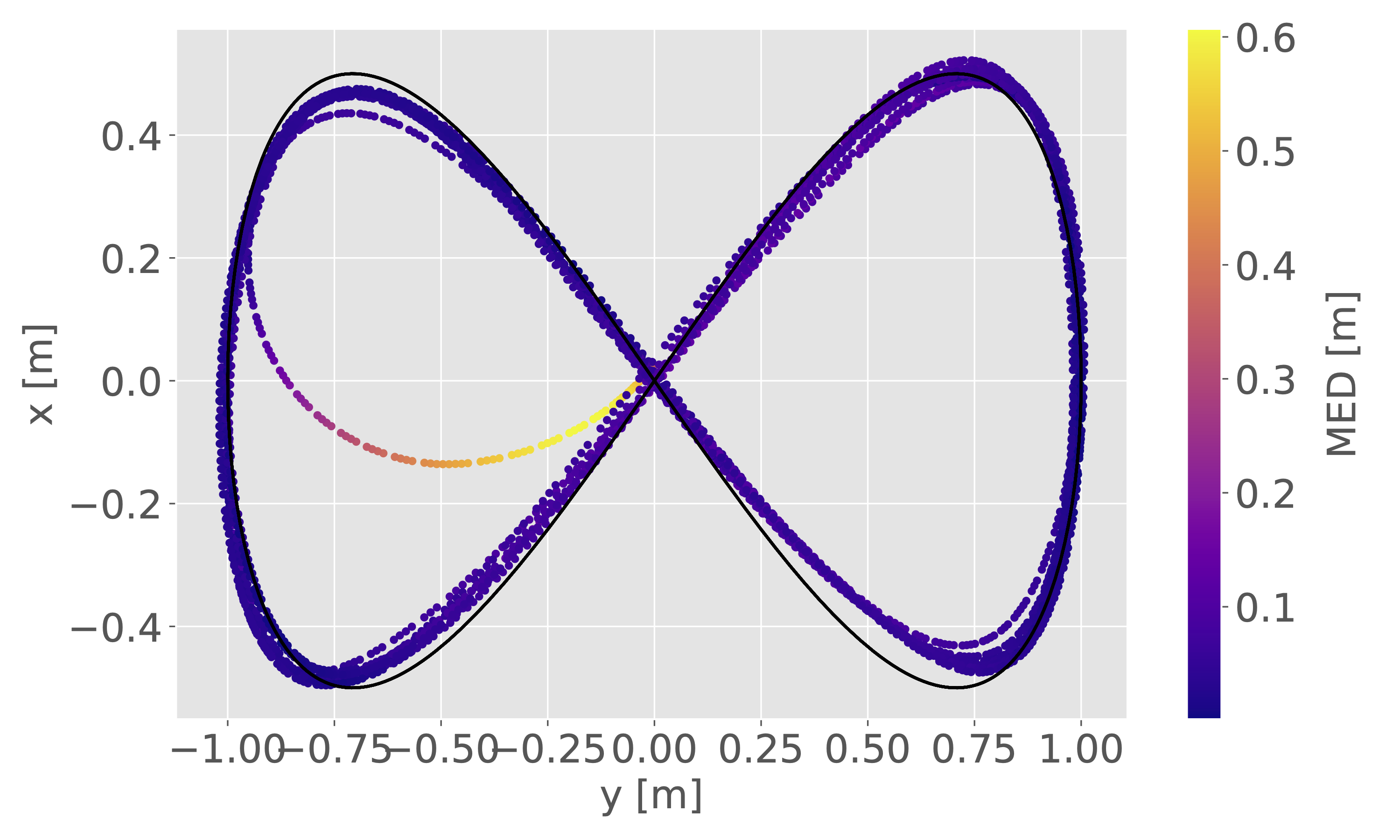}}
\subcaptionbox{Polynomial.}
{\includegraphics[width=0.25\textwidth]{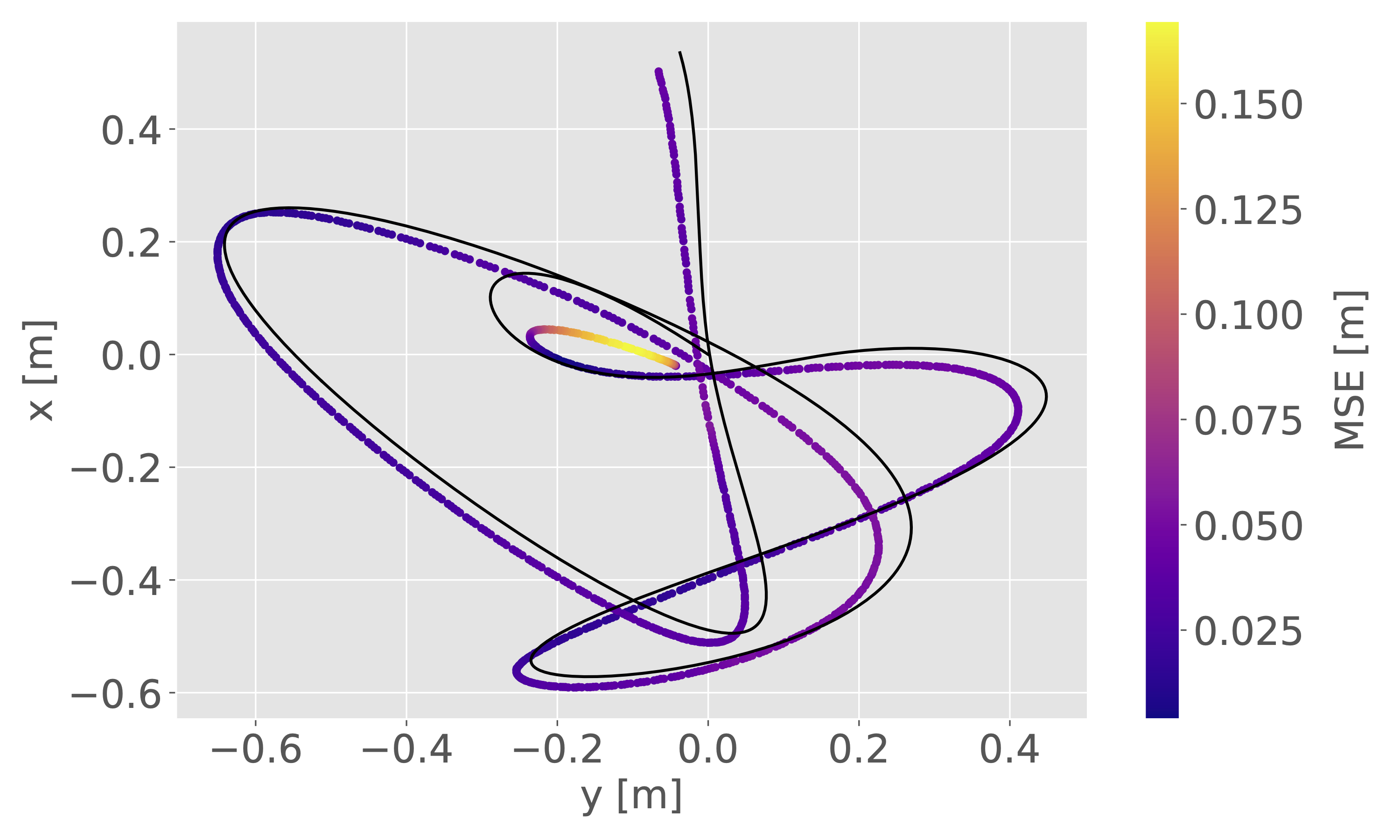}}
\subcaptionbox{Pentagram (fast).}
{\includegraphics[width=0.2\textwidth]{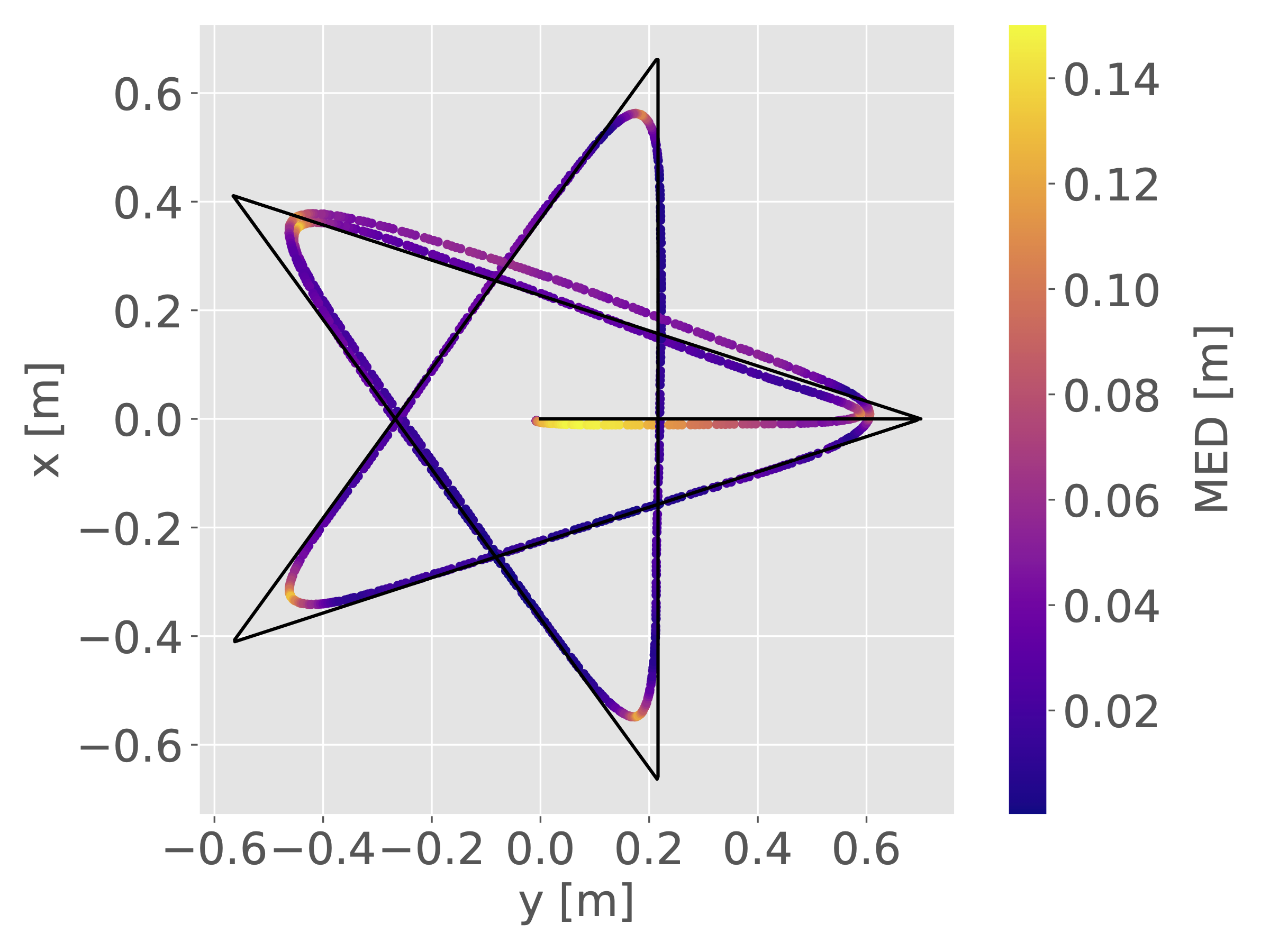}}
\subcaptionbox{Zigzag.}
{\includegraphics[width=0.25\textwidth]{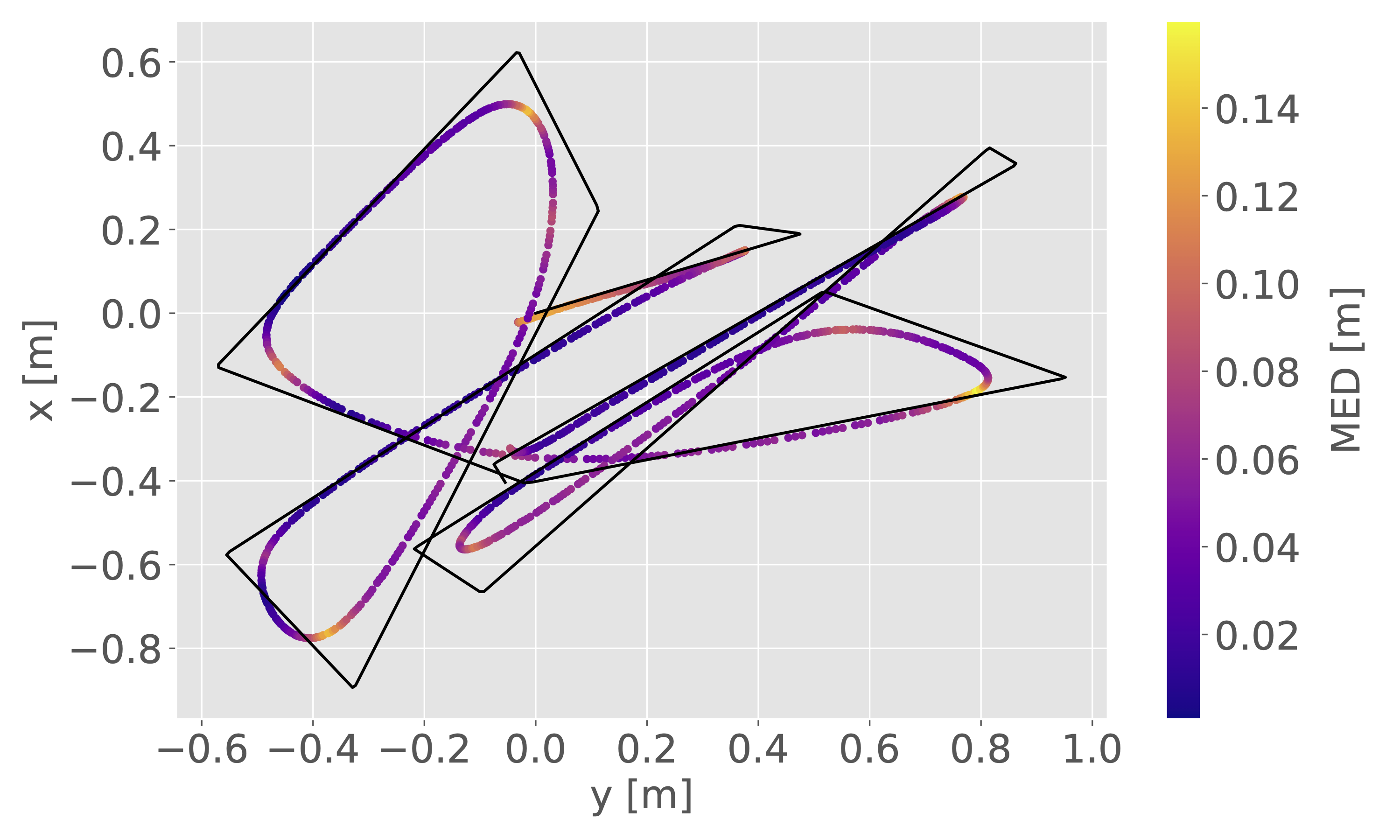}}
\end{minipage}
\caption{Visualization of benchmark trajectories and corresponding trajectories followed using SimpleFlight. The reference trajectories are shown in black.}
\vspace{-6mm}
\label{exp:demo}
\end{figure*}

\subsection{Experiment Setup}

\subsubsection{\textbf{Benchmark Trajectories}}
We adopt two types of trajectories as benchmark trajectories: \textbf{smooth trajectories} (figure-eight and polynomial) and \textbf{infeasible trajectories} (pentagram and zigzag). The figure-eight and pentagram trajectories are deterministic, while the polynomial and zigzag trajectories are randomly generated for each trial. All trajectories start from the origin $(0, 0)$ with a fixed $z$-axis height. Examples of benchmark trajectories are shown in Fig.~\ref{exp:demo}.

\textbf{a. Figure-Eight} 
The figure-eight trajectory is a periodic smooth curve defined as 
$\boldsymbol{p}(t) = [\cos(2\pi t / T), \sin(4\pi t / T) / 2, 1]$,
where $T$ represents the time required to complete one figure-eight lap. We test three velocities by varying $T$: $15.0s$ (Slow), $5.5s$ (Normal), and $3.5s$ (Fast), corresponding to maximum velocities of $0.6$m/s, $1.6$m/s, and $2.5$m/s in the reference trajectories, respectively.

\textbf{b. Polynomial}
The smooth polynomial trajectory consists of multiple randomly generated 5-degree polynomial segments. The duration of each segment is randomly selected between $1.5$s and $4$s. The velocity of the reference trajectories ranges from $0$ to $1$m/s. To ensure smoothness, we enforce continuity of the first, second, and third derivatives at the junctions between consecutive segments.

\textbf{c. Pentagram}
The pentagram is an infeasible trajectory where the quadrotor sequentially visits the five vertices of a pentagram at a constant velocity. We test two different velocities: $0.5$m/s and $1$m/s, marked as Slow and Fast, respectively.

\textbf{d. Zigzag}
The zigzag trajectory is infeasible and is generated based on several randomly selected waypoints, with $x-$ and $y-$ coordinates distributed between $-1$m and $1$m. Consecutive waypoints are connected by straight lines, with time intervals randomly chosen between $1$s and $1.5$s. {The velocity of the reference trajectories ranges from $0$ to $2$m/s.}

We note that pentagram and zigzag trajectories are challenging infeasible due to their sharp directional changes, i.e., infinite acceleration. Successfully tracking these trajectories requires quadrotors to perform long-horizon optimization and operate near the limits of their system performance, which is difficult for quadrotors with low thrust-to-weight ratios.

\subsubsection{\textbf{Training Details}}
We employ OmniDrones~\cite{xu2024omnidrones}, a GPU-accelerated, highly parallel drone simulator, to train the quadrotor control policy using the on-policy PPO algorithm~\cite{schulman2017proximal}. The simulator operates at 100 Hz with a timestep of $0.01$s. {During training, we use a balanced mixture of smooth randomized polynomial trajectories and infeasible zigzag trajectories, generated under consistent benchmark rules. This setup makes figure-eight and pentagram trajectories out-of-distribution (OOD) test cases.} The policy is trained for 15,000 epochs, with a single policy derived for all trajectories.

\subsubsection{\textbf{Evaluation Metric}} 
We assess tracking performance using the Mean Euclidean Distance (MED) between the quadrotor's actual and target positions in the $x$- and $y$-axes, averaged over the entire trajectory. For the figure-eight trajectory, MED is averaged over ten repetitions per trial across three trials. For the pentagram trajectory, MED is computed over three single-run trials. For polynomial and zigzag trajectories, we generate five random trajectories of each type, repeating each twice, resulting in ten trials. MED values are averaged and reported as ``mean (standard deviation)'' in meters (m).

\subsection{In-depth Analysis on Key Factors}\label{sec:ablation}
{We systematically evaluate all proposed key factors through simulation and real-world experiments on figure-eight trajectories. For deployment, we use the Crazyflie 2.1 nano-quadrotor, with position, velocity, and orientation data provided by an OptiTrack motion capture system at 100 Hz. An offboard computer executes the policy, sending CTBR commands to the quadrotor via a 2.4 GHz radio at 100 Hz. Notably, we deploy a single randomly-selected policy (from three training seeds) due to the demonstrated robustness of our training process.}

\subsubsection{\textbf{Factors 1\&2: Input Space Design}}
\begin{figure}[t]
\begin{minipage}{0.49\textwidth}
\centering
\subcaptionbox{Common inputs.\label{fig:a}}
{\includegraphics[width=0.49\textwidth]{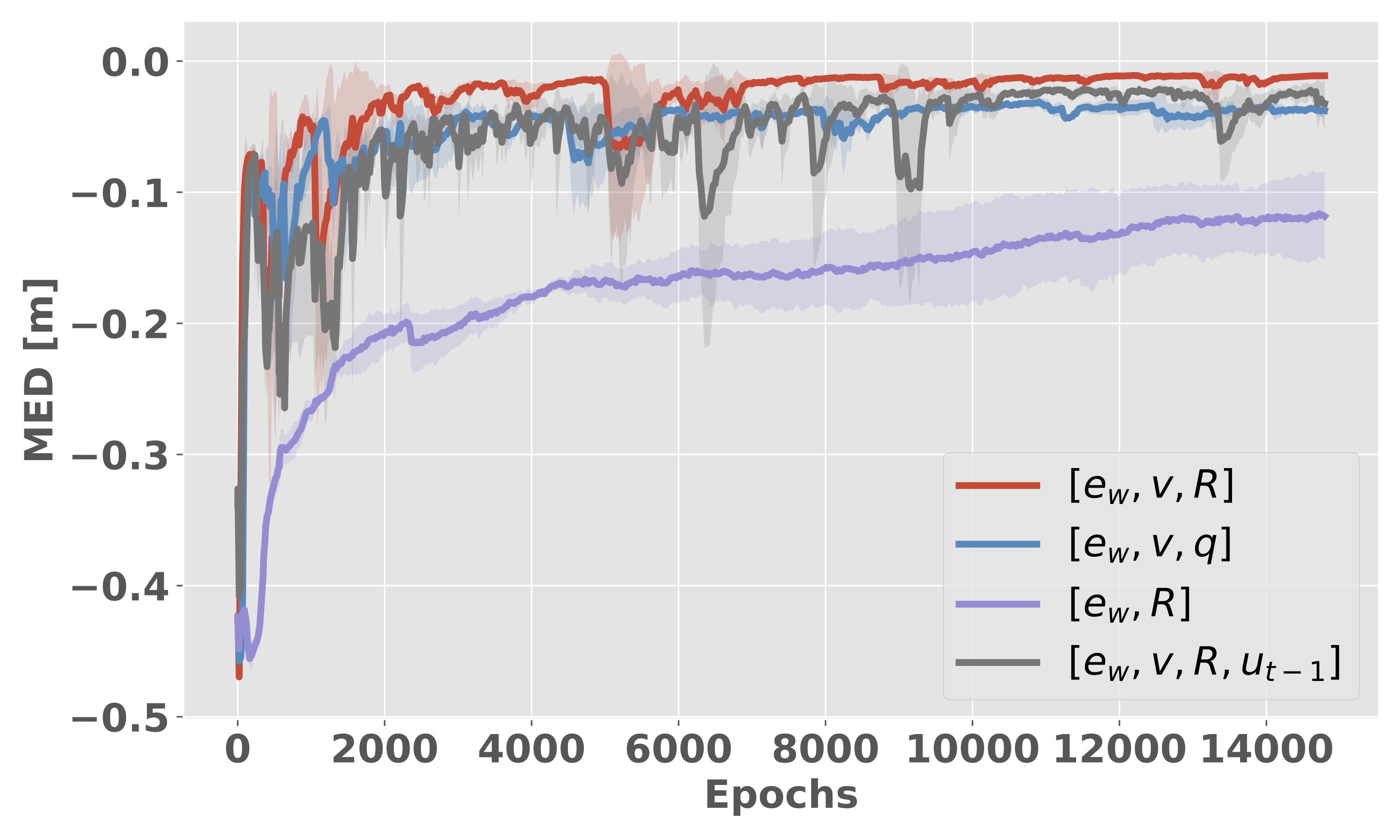}}
\subcaptionbox{Impact of the time vector.\label{fig:c}}
{\includegraphics[width=0.49\textwidth]{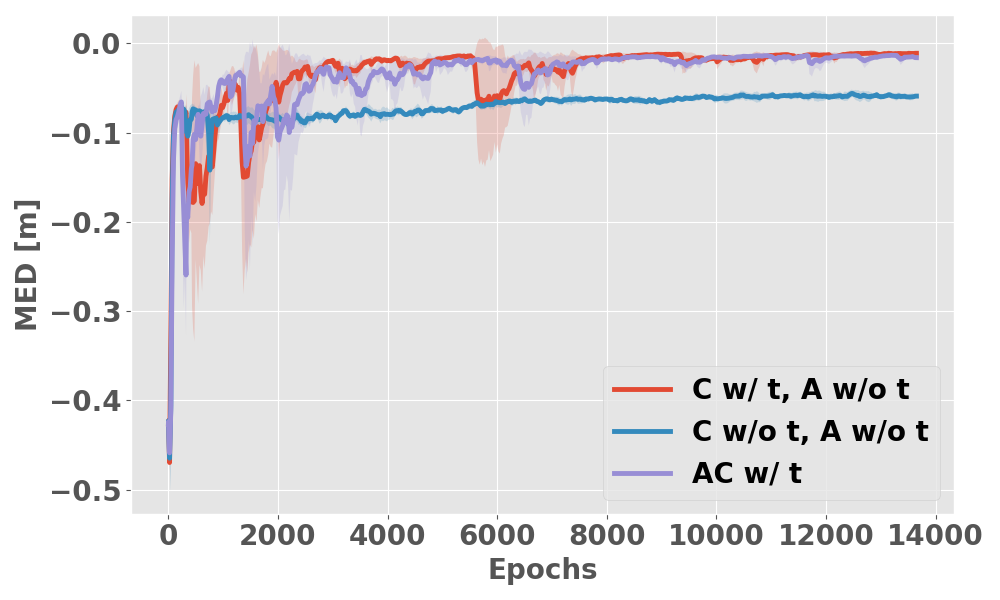}}
\end{minipage}
\begin{minipage}{0.49\textwidth}
\centering
\footnotesize
\vspace{3mm}
\begin{tabular}{cccc}
\toprule
\multirow{3}{*}{Methods}& \multicolumn{3}{c}{Figure-eight} \\ 
\cmidrule{2-4}
& Slow & Normal & Fast \\
\midrule
AC w/ t & 0.003\scriptsize{(0.000)} & 0.010\scriptsize{(0.002)} & 0.033\scriptsize{(0.004)} \\
\midrule
C w/ t, A w/o t & \textbf{0.002\scriptsize{(0.001)}} & \textbf{0.006\scriptsize{(0.001)}} & \textbf{0.024\scriptsize{(0.004)}} \\
\bottomrule
\end{tabular}
\subcaption{The tracking performance on figure-eight with 10 laps.}\label{tab:ten_laps}
\end{minipage}
\caption{Training performance of input space designs. (a) For shared actor-critic inputs, the combination of relative positions $\boldsymbol{e}_{\mathcal{W}}$, linear velocity $\boldsymbol{v}$, and rotation matrix $\boldsymbol{R}$ achieves the lowest simulated MED. (b) Adding the time vector to inputs significantly improves tracking performance.  
(c) Including the time vector in actor inputs slightly degrades performance for long-duration trajectories.}
\vspace{-7mm}
\label{exp:input}
\end{figure}

Fig.~\ref{exp:input} illustrates the training curves for different input space designs. For the common inputs of the actor and critic (Fig.~\ref{fig:a}), $\boldsymbol{q}$ denotes the quaternion corresponding to the rotation matrix $\boldsymbol{R}$. Leveraging the relative positions $\boldsymbol{e}_{\mathcal{W}}$, linear velocity $\boldsymbol{v}$, and the rotation matrix $\boldsymbol{R}$ achieves the best tracking performance in simulation. We observe a training performance degradation {(approximately 63.6\%)} when replacing $\boldsymbol{R}$ with $\boldsymbol{q}$ . This is because representations for the 3D rotations are discontinuous in four or fewer dimensions, making it challenging for neural networks to learn, as also observed in graphics and vision studies~\cite{8953486}.
Excluding the linear velocity $\boldsymbol{v}$ significantly degrades performance, underscoring its importance for predicting actions in agile quadrotor control. Including the previous action $\boldsymbol{u}_{t-1}$, as suggested in prior works~\cite{kaufmann2023champion,dionigi2024power}, results in a slight performance drop. We hypothesize this is because the RL policy continuously evolves during training, and including the previous action in the input introduces non-stationary into the environment, thereby reducing performance.

{Regarding the impact of the time vector (Fig.~\ref{fig:c}), we evaluate three configurations: (1) time vector in both actor and critic (\emph{AC w/ t}), (2) time vector only in critic (\emph{C w/ t, A w/o t}), and (3) no time vector (\emph{AC w/o t}). Results show that incorporating the time vector significantly improves tracking accuracy (\emph{AC w/ t} and \emph{C w/t, A w/o t}), as it enhances the critic's ability to capture temporal information and estimate state values. However, including the time vector in the actor (\emph{AC w/ t}) can cause out-of-distribution (OOD) issues during long-duration flights, as the reference trajectory's timesteps may exceed the training trajectory's maximum length (Tab.~\ref{tab:ten_laps}). Thus, we include the time vector only in the critic to balance accurate value estimation with robust performance.}

\begin{figure*}[t]
\centering
\begin{minipage}{0.24\textwidth}
\centering
\includegraphics[width=\textwidth]{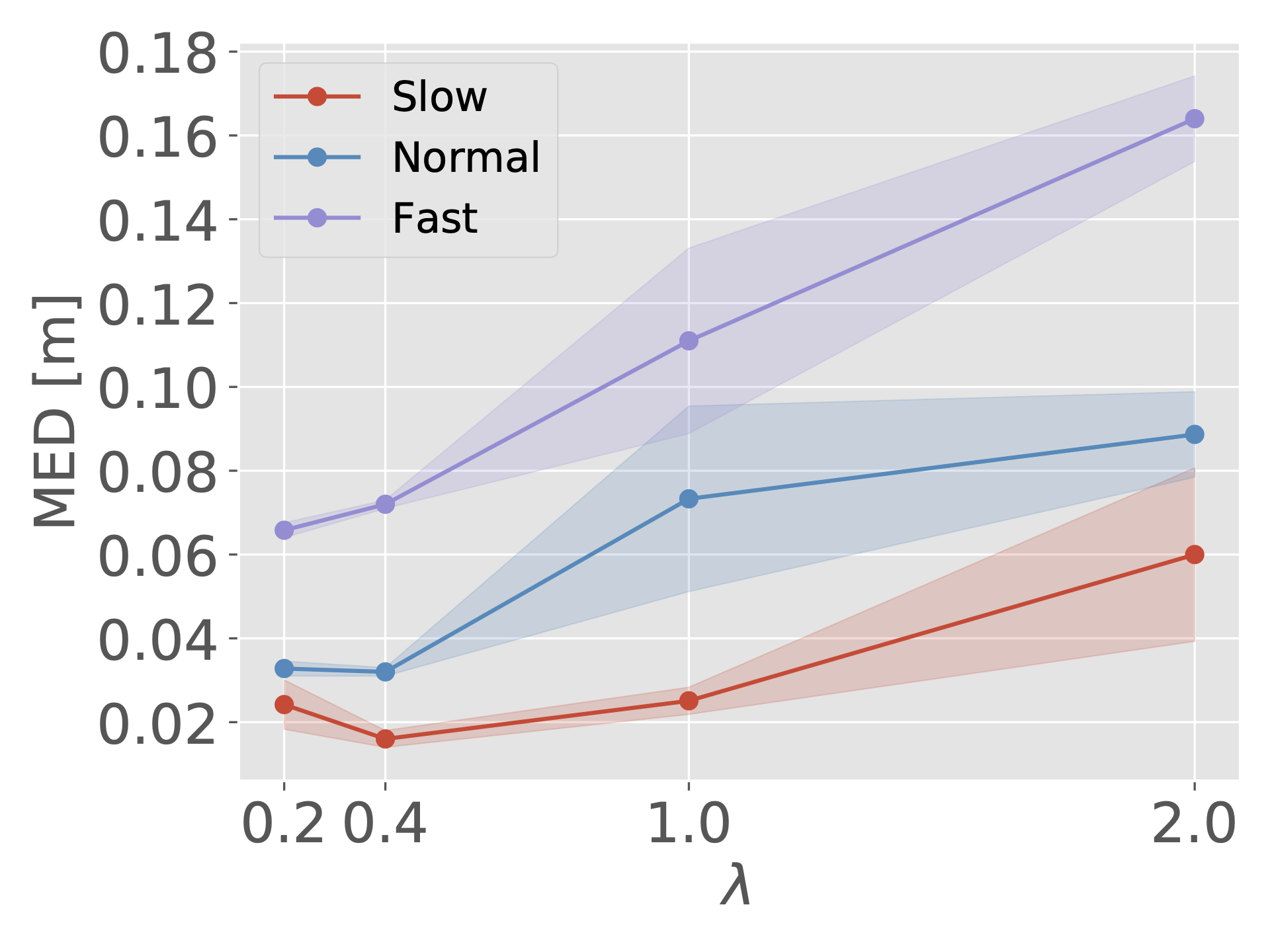}
\caption{Real-world performance of different $\lambda$ on the figure-eight trajectory. We finally choose $\lambda = 0.4$.}
\vspace{-3mm}
\label{exp:lambda}
\end{minipage}
\begin{minipage}{0.72\textwidth}
\centering
\subcaptionbox{Slow.}
{\includegraphics[width=0.32\textwidth]{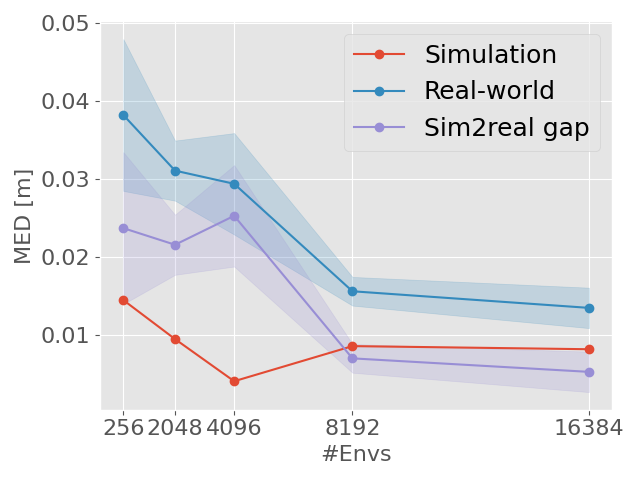}}
\subcaptionbox{Normal.}
{\includegraphics[width=0.32\textwidth]{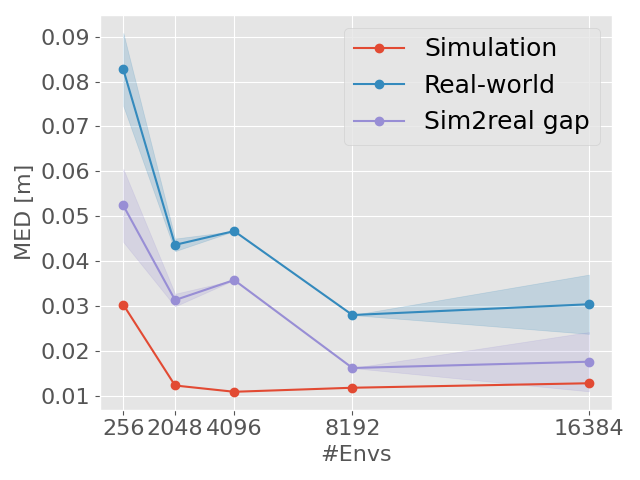}}
\subcaptionbox{Fast.}
{\includegraphics[width=0.32\textwidth]{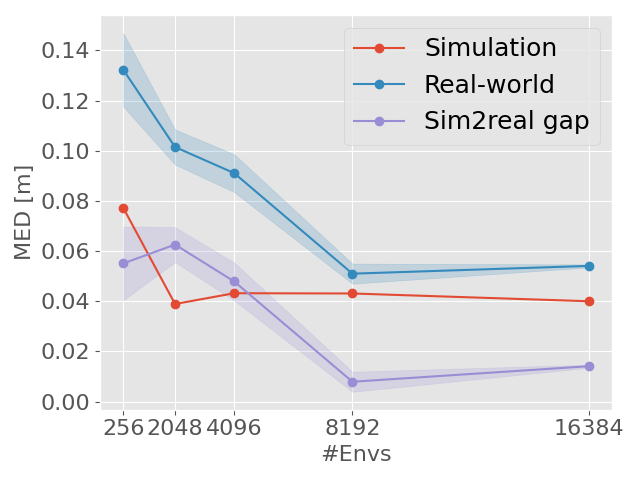}}
\caption{Effect of batch sizes on tracking performance on figure-eight trajectories. Increasing the batch size enhances real-world performance as simulation performance converges, with real-world results also stabilizing as the batch size grows further.}
\vspace{-3mm}
\label{exp:batch}
\end{minipage}
\end{figure*}

\subsubsection{\textbf{Factor 3: Smoothness Reward Design}}
We evaluate various smoothness components commonly used in existing studies, with the real-world tracking performance summarized in Tab.~\ref{tab:smoothness}. Here, $\boldsymbol{acc}_t$,$\boldsymbol{jerk}_t$,$\boldsymbol{snap}_t$ represent the second, third, and fourth derivative of position at timestep $t$, respectively, and $\boldsymbol{u}_t$ denotes the policy's CTBR output at timestep $t$. Note that \( ||\boldsymbol{u}_t||_2 \) penalizes desired angular velocity and thrust, indirectly constraining the third derivative of position, while \( ||\boldsymbol{u}_t - \boldsymbol{u}_{t-1}||_2 \) penalizes angular acceleration and differential thrust, indirectly targeting the fourth derivative. The smoothness reward is defined as $r_{aux} = e^{-A}$, where $A$ represents different forms of smoothness reward components. A grid search is conducted to optimize the hyperparameters for each component, and the best results are reported.
In addition to reward design, we examine two alternative methods to encourage valid and smooth commands: ``Action Clipping (AC)'' and ``Low-Pass Filter (LPF)''. AC directly limits actions exceeding a predefined threshold, while LPF smooths the policy output as $ \boldsymbol{u}_t' = \alpha \boldsymbol{u}_t + (1 - \alpha) \boldsymbol{u}_{t-1} $, where $\alpha$ is the cutoff parameter. Among these, \( ||\boldsymbol{u}_t - \boldsymbol{u}_{t-1}||_2 \) achieves the best real-world tracking performance. 
In contrast, non-reward methods like LPF fail to generalize across velocities, and AC limits agile maneuvers, resulting in suboptimal performance for high-velocity trajectories. 
Direct action constraints such as \( ||\boldsymbol{u}_t||_2 \) outperform indirect kinematic constraints like \( ||\boldsymbol{jerk}_t||_2 \) on fast trajectories, highlighting the effectiveness of action-level regularization for challenging tasks.
\begin{table}[htp]
\vspace{-2mm}
\centering
\footnotesize
{
{\begin{tabular}{cccc} 
\toprule
 &Slow&Normal &Fast \\ 
\midrule
 Action Clipping & 0.035\scriptsize{(0.001)} & 0.077\scriptsize{(0.016)} & 0.310\scriptsize{(0.016)} \\ 
\midrule
                     Low-Pass Filter & $\infty$ & $\infty$ & $\infty$ \\ 
\midrule
$||\boldsymbol{acc}_t||_2$& 0.183\scriptsize{(0.004)} &0.329\scriptsize{(0.004)} & $\infty$ \\ 
\midrule
                     $||\boldsymbol{jerk}_t||_2$& 0.024\scriptsize{(0.009)} & 0.047\scriptsize{(0.005)} & $\infty$ \\ 
\midrule
                     $||\boldsymbol{snap}_t||_2$& 0.026\scriptsize{(0.002)} & $\infty$ & $\infty$ \\ 
\midrule
                     $||\boldsymbol{u}_t||_2$& 0.044\scriptsize{(0.003)} & 0.066\scriptsize{(0.002)} &  0.110\scriptsize{(0.027)} \\ 
\midrule
                     $||\boldsymbol{u}_t - \boldsymbol{u}_{t-1}||_2$& \textbf{0.016\scriptsize{(0.002)}} & \textbf{0.028\scriptsize{(0.000)}} &  \textbf{0.051\scriptsize{(0.002)}} \\ 
\bottomrule
\end{tabular}}}
\caption{Real-world tracking performance of different smoothness components that encourage valid and smooth control commands. The action-level regularization $||\boldsymbol{u}_t - \boldsymbol{u}_{t-1}||_2$ achieves the best tracking performance among all designs.}
\vspace{-3mm}
\label{tab:smoothness}
\end{table}

Since there exists a trade-off between the task reward and the smoothness reward, we conduct experiments on the smoothness reward coefficient $\lambda$, with the real-world tracking performance shown in Fig.~\ref{exp:lambda}. For $\lambda = 0.2$, not all fast trajectory experiments are successful; only results from successful trials are reported. As observed, larger $\lambda$ can degrade tracking performance due to restricted agility, while smaller $\lambda$ may lead to unstable flight. Based on these findings, we set $\lambda = 0.4$ for a trade-off.

\begin{table*}[t]
\centering

\begin{tabular}{cccccc} 
\toprule
      & Offset$+30\%$      & Offset+DR$30\%$ & SysID+DR$30\%$ & SysID+DR$10\%$ & SysID                              \\ 
\midrule
Mass $m$                  & $\infty$ &  $\infty$&0.066\scriptsize{(0.007)} & 0.041\scriptsize{(0.006)}&\multirow{5}{*}{\textbf{\textbf{0.028\scriptsize{(0.000)}}}}  \\ 
\cmidrule{1-5}
Inertia $\boldsymbol{I}$  & 0.041\scriptsize{(0.004)} &0.046\scriptsize{(0.002)}  &0.053\scriptsize{(0.005)} & 0.036\scriptsize{(0.001)} &                                            \\ 
\cmidrule{1-5}
Motor Time Constant $T_m$       & 0.036\scriptsize{(0.002)} & 0.044\scriptsize{(0.002)} & 0.057\scriptsize{(0.012)} &0.040\scriptsize{(0.001)}  &                                           \\ 
\cmidrule{1-5}
Thrust Coefficient $k_f$  & 0.107\scriptsize{(0.013)}  & 0.035\scriptsize{(0.001)}  & 0.050\scriptsize{(0.005)} & 0.034\scriptsize{(0.002)}&                                          \\
\bottomrule
\end{tabular}
\caption{Real-world tracking performance of SysID and DR on the figure-eight trajectory at normal velocity. For sensitive parameters, such as the thrust coefficient $k_f$, when accurate calibration is not feasible, DR can effectively enhance the robustness of the policy. For non-sensitive parameters, DR primarily increases the learning difficulty without providing significant benefits.
}
\label{tab:dynamics}
\end{table*}

\begin{table*}[htp]
\centering
\footnotesize
{
{\begin{tabular}{ccccccccc} 
\toprule
 \multirow{3}{*}{Platform}&\multirow{3}{*}{Methods}&\multicolumn{3}{c}{Figure-eight}&\multirow{3}{*}{Polynomial} &\multicolumn{2}{c}{Pentagram}&\multirow{3}{*}{Zigzag} \\ 
 \cmidrule{3-5}
 \cmidrule{7-8}
& &Slow&Normal&Fast&&Slow&Fast& \\
 \midrule
\multirow{4}{*}{Crazyflie}&Fly~\cite{eschmann2024learning} & 0.093\scriptsize{(0.001)} & 0.181\scriptsize{(0.004)} & 0.282\scriptsize{(0.012)} & 0.289\scriptsize{(0.042)} &0.104\scriptsize{(0.005)} &$\infty$& $\infty$ \\ 
\cmidrule{2-9}
& DATT~\cite{huang2023datt} & 0.050\scriptsize{(0.009)} & $\infty$ & $\infty$ & 0.081\scriptsize{(0.019)} &0.055\scriptsize{(0.004)} &0.146\scriptsize{(0.012)}& 0.114\scriptsize{(0.019)}* \\ 
\cmidrule{2-9}
&  SimpleFlight (50Hz) & 0.035\scriptsize{(0.004)} & 0.056\scriptsize{(0.003)} & 0.114\scriptsize{(0.006)} & 0.042\scriptsize{(0.007)} &0.031\scriptsize{(0.003)} &\textbf{0.043\scriptsize{(0.000)}} & 0.053\scriptsize{(0.009)}  \\
\cmidrule{2-9}
&  SimpleFlight (100Hz) & \textbf{0.016\scriptsize{(0.002)}} & \textbf{0.028\scriptsize{(0.000)}} & \textbf{0.051\scriptsize{(0.002)}} & \textbf{0.032\scriptsize{(0.003)}} &\textbf{0.024\scriptsize{(0.001)}} &0.045\scriptsize{(0.002)} & \textbf{0.052\scriptsize{(0.003)}}  \\ 
\midrule
\multirow{3}{*}{Air}&PAMPC~\cite{falanga2018pampc} & \textbf{0.011\scriptsize{(0.001)}} & 0.051\scriptsize{(0.009)} & 0.117\scriptsize{(0.001)} & \textbf{0.028\scriptsize{(0.006)}} & \textbf{0.017\scriptsize{(0.000)}} & 0.051\scriptsize{(0.002)} & 0.064\scriptsize{(0.007)} \\
\cmidrule{2-9}
&SimpleFlight (Ours)  & 0.014\scriptsize{(0.001)} & \textbf{0.036\scriptsize{(0.000)}} & \textbf{0.077\scriptsize{(0.001)}} & \textbf{0.028\scriptsize{(0.007)}} & 0.019\scriptsize{(0.000)} & \textbf{0.042\scriptsize{(0.001)}} & \textbf{0.046\scriptsize{(0.007)}} \\
\bottomrule
\end{tabular}}}
\caption{SimpleFlight demonstrates superior real-world performance across benchmark trajectories. On the Crazyflie platform, it reduces MED by over 50\% compared to all baselines. On the Air platform, SimpleFlight shows comparable performance to Crazyflie results while outperforming finely-tuned PAMPC, confirming cross-platform generalization. * indicates that for DATT in the zigzag trajectory trials, 4 out of 10 attempts failed; the reported MED reflects the 4 successful trials.}
\label{tab:main}
\vspace{-5mm}
\end{table*}

\subsubsection{\textbf{Factor 4: SysID and DR}}
We analyze the effects of SysID and DR on tracking performance for figure-eight trajectories at normal velocity in Tab.~\ref{tab:dynamics}. 
We begin by calibrating four key dynamic parameters—mass $m$, inertia $\boldsymbol{I}$, motor time constant $T_m$, and thrust coefficient $k_f$—and report results using these calibrated values as ``SysID''. We then apply DR to each parameter, exploring two parameter randomization ranges: $[-10\%, +10\%]$ and $[-30\%, +30\%]$, denoted as ``SysID+DR$10\%$'' and ``SysID+DR$30\%$,'' respectively. Our findings indicate that applying DR to approximately well-calibrated parameters generally does not improve sim-to-real transfer performance as it may increase learning complexity. In addition, introducing DR leads to slightly higher standard deviations, indicating that it can induce more unstable behavior. To further simulate inaccurate calibration, we introduce a $+30\%$ offset to each dynamics parameter, referred to as ``Offset+$30\%$.'' {The results reveal varying performance sensitivity across parameters. Precise measurement is critical for sensitive parameters such as mass $m$ and thrust coefficient $k_f$, while strict calibration is less essential for less sensitive parameters like inertia $\boldsymbol{I}$ and motor time constant $T_m$.}
We then apply DR to these offset values (termed "Offset+DR$30\%$") and observe a noticeable improvement in the shifted thrust coefficient’s performance, alongside comparable results for the other parameters. {Therefore, for sensitive parameters, such as the thrust coefficient $k_f$, when accurate calibration is not feasible, DR can effectively enhance the robustness of the policy. For non-sensitive parameters, DR primarily increases the learning difficulty without providing performance benefits.}

{
\subsubsection{\textbf{Factor 5: Effect of Batch Sizes}}
To evaluate the impact of the batch sizes, we test simulation and real-world performance using figure-eight trajectories (slow, normal and fast) via varying parallel environments. As shown in Fig.~\ref{exp:batch}, increasing the batch size enhances real-world performance as simulation performance converges, with real-world results also stabilizing as the batch size grows further. Based on this finding, we recommend using larger batch sizes during training to enhance sim-to-real transfer.}

{
\subsection{Comparison of other methods}
\subsubsection{\textbf{Baselines}}

While our paper focuses on robust RL policies for zero-shot real-world deployment (i.e. relative performance), we also show SimpleFlight’s strong absolute performance. We benchmark two SOTA RL methods (DATT~\cite{huang2023datt} and Fly~\cite{eschmann2024learning}) on the Crazyflie and stress-test SimpleFlight against a well-tuned  MPC method, PAMPC~\cite{falanga2018pampc} on our custom quadrotor, named Air, which features a 250mm arm length and is equipped with a PX4 flight controller and an Nvidia Orin processor.

\textbf{a. DATT}~\cite{huang2023datt} is a feedforward-feedback-adaptive policy for CTBR command-based trajectory tracking, achieving SOTA performance over PID and non-linear MPC~\cite{williams2017information}. We retrain DATT on the standard Crazyflie 2.1 (body rate: $[-\pi, \pi]\text{rad/s}$, acceleration: $[0, 1.6g]$) to ensure fair comparison, retaining its disturbance estimation for optimal performance. Deployment follows the same protocol as SimpleFlight.

\textbf{b. Fly}~\cite{eschmann2024learning} proposes a high-speed simulator and RL-based framework for direct RPM control, enabling superior sim-to-real transfer. Using the released checkpoint, we perform onboard inference to evaluate its performance on benchmark trajectories, as it outperforms PID and prior RL policies.

{\textbf{c. PAMPC}~\cite{falanga2018pampc} is a non-linear MPC method that jointly optimizes perception and action objectives. By excluding the vision objective and fine-tuning cost function hyperparameters, we adapt it to ensure accurate tracking performance across all benchmark trajectories.}

\subsubsection{\textbf{Real-world experiments}}
We first report the trajectory tracking performance of SimpleFlight compared to the baseline methods across all benchmark trajectories on the Crazyflie, as shown in Tab.~\ref{tab:main}. SimpleFlight significantly outperforms all baselines across benchmark trajectories on the Crazyflie (Tab.~\ref{tab:main}), reducing MED by over 50\% and achieving the lowest standard deviation, indicating superior stability. Fly~\cite{eschmann2024learning} reliably tracks smooth trajectories at varying velocities but struggles with infeasible paths (e.g., fast pentagram and zigzag) due to limited long-horizon reasoning. DATT~\cite{huang2023datt} handles infeasible trajectories aggressively but fails in high-velocity tracking on low thrust-to-weight quadrotors. SimpleFlight excels in actuation constraint awareness, long-horizon reasoning, and optimization, particularly for sharp turns and complex maneuvers. At 50 Hz, SimpleFlight shows a minor performance drop compared to 100 Hz but still surpasses DATT, as higher-frequency control enables faster error correction.

We also conduct experiments on our own quadrotor platform, Air, to validate the effectiveness of SimpleFlight. The results demonstrate that our method achieves comparable performance on the Air platform to that on the Crazyflie, slightly outperforming the finely tuned PAMPC. This highlights SimpleFlight's ability to generalize across quadrotor models and sizes. We also provide flight videos on the Air platform on our website.

We remark that the comparison in Tab.~\ref{tab:main} may not be entirely fair, as the policies are trained using different simulators, modeling approaches, and inputs/outputs. What we aim to convey here is that SimpleFlight, to the best of our knowledge, achieves the best control performance, despite not incorporating any algorithmic or architectural improvements. As a collection of proposed key factors, SimpleFlight can be integrated on top of existing quadrotor control methods. 
}

\section{CONCLUSION}\label{sec:conclusion}
\textbf{SimpleFlight} is an RL framework for robust zero-shot deployment of quadrotor control policies. By incorporating five key factors including enhanced inputs design with velocity and rotation matrix, time vector for critic, regularization of the difference between successive actions as a smoothness reward, selected domain randomization with system identification, and large training batch sizes, it effectively bridges the sim-to-real gap. Evaluations on a Crazyflie quadrotor demonstrate over 50\% lower tracking error compared to SOTA RL baselines. 


\bibliographystyle{IEEEtran}

\end{document}